\documentclass[runningheads]{llncs}
\usepackage[T1]{fontenc}
\usepackage{makecell}
\usepackage{amsmath}
\usepackage{graphicx}
\usepackage{amssymb}

\DeclareMathAlphabet\mathbfcal{OMS}{cmsy}{b}{n}

\usepackage{hyperref}
\usepackage{caption}
\usepackage{subcaption}
\usepackage{comment}
\usepackage{amsmath,amssymb,amsfonts}
\usepackage{caption}
\usepackage{subcaption}
\usepackage{multirow}
\usepackage{algorithmic}
\usepackage{graphicx}
\usepackage{float}
\usepackage{textcomp}
\usepackage{booktabs}
\usepackage{tabularx}
\usepackage{cleveref}
\usepackage{comment}
\usepackage{xcolor}

\usepackage{cite}
\usepackage{amsmath}
\usepackage{algorithmic}
\usepackage{graphicx}
\usepackage{float}
\usepackage{caption}
\usepackage{subcaption}
\usepackage{textcomp}
\usepackage{booktabs}
\usepackage{multirow}
\usepackage{caption}
\usepackage{tabularx}
\usepackage{cleveref}

\begin{document}

\title{Deep BI-RADS Network for Improved Cancer Detection from Mammograms}
\titlerunning{Deep BI-RADS}
%
\author{Gil Ben-Artzi\inst{1,2} \and Feras Daragma\inst{1,3} \and Shahar Mahpod\inst{1,2} }
\authorrunning{G. Ben-Artzi et al.}
%
\institute{School of Computer Science, Ariel University, Ariel, Israel \and
\email{\{gilba,mahpods\}@ariel.ac.il}\\ \and
\email{daragma.feras@gmail.com}}

\maketitle

\begin{abstract}
While state-of-the-art models for breast cancer detection leverage multi-view mammograms for enhanced diagnostic accuracy, they often focus solely on visual mammography data. However, radiologists document valuable lesion descriptors that contain additional information that can enhance mammography-based breast cancer screening. A key question is whether deep learning models can benefit from these expert-derived features. To address this question, we introduce a novel multi-modal approach that combines textual BI-RADS lesion descriptors with visual mammogram content. Our method employs iterative attention layers to effectively fuse these different modalities, significantly improving classification performance over image-only models. Experiments on the CBIS-DDSM dataset demonstrate substantial improvements across all metrics, demonstrating the contribution of handcrafted features to end-to-end.
\end{abstract}


\section{Introduction}
\label{sec:intro}

In recent years, deep learning techniques have emerged as a powerful tool for breast cancer detection, demonstrating significant potential in enhancing the accuracy of mammography interpretation. State-of-the-art models~\cite{Aguilar2020,nguyen2022novel,yan2021towards} have achieved impressive results by leveraging information from different mammogram views (craniocaudal (CC) and mediolateral oblique (MLO)) to enhance diagnostic accuracy. However, these approaches often focus solely on end-to-end extracted visual features.

Radiologists use the Breast Imaging Reporting and Data System (BI-RADS) lexicon~\cite{ACR-BIRADS} to document specific lesion descriptors such as size, shape, and margin characteristics during mammogram interpretation. These descriptors can offer crucial insights that aid in distinguishing between benign and malignant lesions. In this paper we investigate whether incorporating BI-RADS descriptors can improve deep learning for cancer detection. 

Integrating these descriptors with mammograms poses challenges due to differences in modalities, scales, importance levels, and inconsistencies across radiology reports. To address these challenges and answer our research question, we propose a multi-modal dual-branch architecture. Each branch, corresponding to CC/MLO views, encodes the mammogram in a multi-resolution manner. We introduce a dedicated iterative attention mechanism~\cite{jaegle2021perceiver} that processes input from the previous layer, the current encoded resolution of the mammogram, and processed information from the other branch. By processing information from these sources at each level using the attention mechanism, our model effectively overcomes the differences in modalities and inconsistencies.

We conduct experiments using the CBIS-DDSM dataset, which includes both mammograms and BI-RADS descriptors as metadata. Our results indicate that our multi-modal iterative attention-based approach effectively integrates both visual and textual modalities, outperforming image-only models for benign vs. malignant classification. We achieve performance improvements across all metrics compared to image-only models, with an AUC score of 0.87. Our results demonstrate the significant potential of incorporating handcrafted features with deep learning models, suggesting a promising direction for future research in medical image analysis.

\section{Related Work}

\subsection{Handcrafted Features for Cancer Detection}

The Breast Imaging Reporting and Data System (BI-RADS)~\cite{ACR-BIRADS}, developed by the American College of Radiology (ACR), acts as a standardized language for describing and classifying breast lesions identified through mammograms, ultrasounds, and MRIs. This system plays a crucial role in improving the consistency, clarity, and accuracy of breast imaging reports. Unlike our suggested features, BI-RADS descriptors are based on the grayscale level of the pixels in the lesions. A similar lexicon, the Thyroid Imaging Reporting and Data System (TI-RADS), has been proposed for thyroid lesions \cite{ACR-TIRADS}.

\subsection{Multi-View Cancer Detection}
Liu et al.~\cite{liu2020cross} presented a cross-view correspondence reasoning method based on a bipartite graph convolutional network for mammogram mass detection. This approach effectively addresses the challenge of inherent view alignment between different views by learning geometric constraints. Tulder et al. \cite{tulder2021cross} proposed a multi-view analysis method for unregistered medical images using cross-view transformers, addressing the challenge of effectively combining features from unregistered mammogram views (CC/MLO) with perspective differences. Shen et al.~\cite{shen2021interpretable} presented an interpretable classifier for high-resolution breast cancer screening images utilizing weakly supervised localization. This approach effectively addresses the challenge of interpretability in deep learning models for mammogram analysis. Chen et al.~\cite{chen2022multi} proposed a multi-view local co-occurrence and global consistency learning method for mammogram classification generalization, addressing the challenge of effectively combining features from unregistered mammogram views (CC/MLO) with perspective differences. While these methods address multi-view analysis, they do not utilize the textual lesion attribute data and cross-view information at each analysis stage - key capabilities of our architecture.

\subsection{Incorporating Handcarfted Features}

In the field of mammogram-based deep learning for breast cancer detection, current research primarily focuses on predicting BI-RADS descriptors as model outputs. The integration of both these descriptors and visual features in mammogram analysis remains an open research question.

Zhang et al. introduced BI-RADS-NET~\cite{zhang2021bi}, an explainable deep learning approach for breast cancer diagnosis that outputs BI-RADS descriptors to better explain predictions, although their model was designed for ultrasound images. ~\cite{9871564,tsai2022high} investigated a deep learning method that utilizes multi-view mammogram images to enhance BI-RADS and breast density assessment, rather than integrating them as in our approach. Liu et al.~\cite{liu2021deep} explored the potential of combining mammography-based deep learning with clinical factors such as age and family history of breast cancer, demonstrating the potential benefits of integrating additional features with visual data in the prediction process.

\begin{figure*}[!t]
\centerline{\includegraphics[width=1\columnwidth]{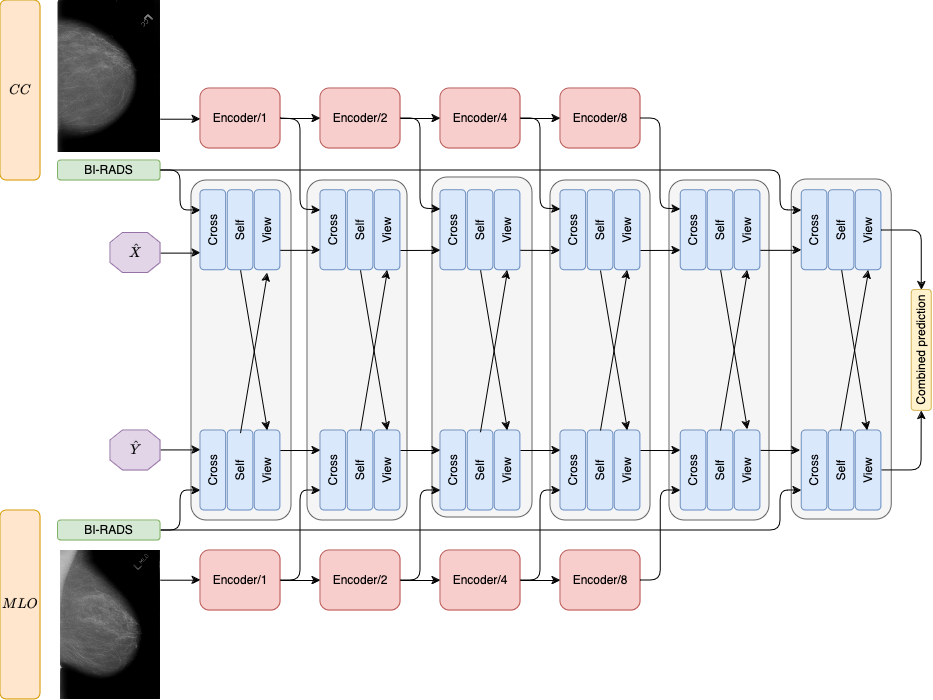}}
\caption{Our model takes mammograms from the MLO and CC views along with a varying number of textual descriptors classes describing one or more lesions as input. The multi-attention layers (grayed blocks) processes these descriptors along with visual features extracted from mammogram images in different resolutions.}
\centering
\label{fig:model}
\end{figure*}

\begin{table}[tb]
\centering
\caption{The BI-RADS descriptors and descriptors classes in the CBIS-DDSM dataset.}
\renewcommand{\arraystretch}{1.2}
\begin{tabular}{lcc}
\toprule
 \textbf{Mass} & & \textbf{Calcifications} \\
\midrule
\underline{\textbf{Margin}} & & \underline{\textbf{Morphology}}\\
Circumscribed & & Pleomorphic \\
Ill-defined & & Amorphous \\
Spicular & & Linear \\
Obscured & & Punctate \\
\midrule
\underline{\textbf{Shape}} & & \underline{\textbf{Distribution}}\\
Round & & Clustered \\
Oval & & Scattered \\
Irregular & & Diffuse \\
\bottomrule
\end{tabular}
\label{table:attributes_examples}
\end{table}

\section{Model Architecture}

Our model consists of two branches. Each branch is composed of $N=6$ stacked identical attention-based layers. An overview of our dual-branch architecture using stacked multi-attention layers (gray background) is presented in Fig. \ref{fig:model}.

The attention based layers progressively fuse and process the multi-modal inputs. The input to the first layer is textual attributes with a skip connection to the last layer. In the first layer, learnable query vectors $\hat{X}$ and $\hat{Y}$ are used since no feature queries exist yet. The input to subsequent layers is the extracted image features at different resolutions using the Big Transfer (BiT) blocks~\cite{kolesnikov2020big}, the input from the preceding layer, and the latent features from the other branch.

The output of the final attention layer in each branch is aggregated by averaging to obtain a unified vector $z \in \mathbb{R}^{L\times 1}$. This representation encodes the joint contribution of images and text. The vector is then layer-normalized and reduced into a labeling vector using a fully-connected layer for the benign and malignant classes.

In the following we present a detailed description of each component of our model. 

\subsection{BI-RADS Descriptors Encoding}
\label{subsec:attributes}
 
Our model utilizes the textual metadata associated with each mammogram, which contains the classes of the lesion and breast descriptors of the Breast Imaging Reporting and Data System (BI-RADS) lexicon~\cite{ACR-BIRADS}. We do not use subjective assessments reflecting radiologist suspicion, like BI-RADS scores, but only the descriptive physical lesion and breast characteristics annotated during routine screening. \Cref{table:attributes_examples} presents examples of the descriptors and descriptors classes that are incorporated by our model. Both calcifications and mass lesions can have combinations of these descriptors. For instance, a mass lesion can have a "Circumscribed-Obscured" margin or a "Round-Oval" shape. Our approach allows for the integration of a variable number of classes, as well as their combinations. 

We assign a unique index $i=1,\dots,N$ to enumerate the possible values of the descriptors classes across all categories. The input to our model is a binary vector $\boldsymbol{\phi} \in \mathbb{R}^L$ (we use $L=256$), defined as:

\begin{equation}
\boldsymbol{\phi}(i) = \begin{cases}
1, & \text{if descriptor class $i$ exists in the description} \\
0, & \text{otherwise}.
\end{cases}
\end{equation}

The encoded input vector $\boldsymbol{\phi}$ represents the BI-RADS descriptors for a single lesion. However, there can be cases with more than one lesion. Our architecture supports a dynamic number of input vectors, so the input is $\Phi = \{\boldsymbol{\phi}_j\}_{j=1}^K$ where $K$ is the number of lesions in the mammogram.

\subsection{Feature Extraction}
\label{subsec:fe}

We use BiT layers as our feature extractor, pre-trained on the PatchCamelyon dataset \cite{Veeling2018-qh}. Our BiT layer $\mathbf{F}$ is based on ResNet50-V2\cite{he2016deep,he2016identity} with modifications made by \cite{kolesnikov2020big} to Group Normalization \cite{DBLP:conf/eccv/WuH18} instead of Layer Normalization\cite{DBLP:journals/corr/BaKH16}, and the use of Weight Standardization~\cite{journals/corr/abs-1903-10520} for all convolution layers. The output of each of these blocks is the input to our multi-attention layer. Each BiT layer $\mathbfcal{F}$ reduces the resolution and increases the number of channels using the following formulation:

\begin{equation}
        \begin{aligned}[b]
          \mathbf{F}_0   &= \mathbfcal{F}_0(I)  \\
           \mathbf{F}_k   &= \mathbfcal{F}_k(\mathbf{F}_{k-1}), \quad \forall \quad k=1,\ldots,N-1, 
        \end{aligned}
\label{eq:features_eq1}
\end{equation}

$I \in \mathbb{R}^{1 \times H \times W}$ is the input image of height $H$ and width $W$, $\mathbf{F}_k \in \mathbb{R}^{d_0 \cdot 2 ^k  \times H' \times W'}$  where $H' = \frac{H}{4\cdot 2^k}$ ,  $W'= \frac{W}{4\cdot 2^k}$ and $d_0=64$.

\subsection{Multi-Attention Layer}

The multi-attention layer has three attention-based~\cite{vaswani2017attention} sub-layers. The first is a cross-attention mechanism, the second is self-attention and the third is view-attention. They enable the model to establish connections between different resolutions and the attributes, between patches within the same image, and between images from different views. 

The utilization of attention enables the exploration of connections between a provided query $Q$, pre-existing key data $K$, while representing these relationships using $V$. It is stated as follows:

\begin{equation}
    \mathbfcal{A}\mathbf{ttn}(Q,K,V) = Softmax(\frac{Q K^{T}}{\sqrt{d}})V,
    \label{eq:attention}
\end{equation}

where $d$ is the scaling factor corresponding to the dimensionality of the key vectors.

\subsubsection{Cross-Attention} 

The first sub-layer, referred to as cross-attention, allows efficient processing of multi-modal inputs including attributes, latent features and images, without relying on domain-specific assumptions. It takes a high-dimensional input and projects it into a lower-dimensional latent bottleneck~\cite{jaegle2021perceiver}. It then applies Transformer-style self-attention on this latent space. It combines the preceding latent features with either the attributes or image features at a given resolution.

The cross-attention in layer $k$, denoted as $\mathbf{A}^C_{k}$, is defined as:

\begin{equation}
\mathbf{A}^C_k:=\mathbfcal{A}\mathbf{ttn}(\mathbf{C}_{k-1},\mathbf{F}_{k-1},\mathbf{F}_{k-1}),
\label{eq:cross_attention}
\end{equation}

where $\mathbf{F}_{k-1}$ is the extracted features from the previous layer (\Cref{subsec:fe}) and $\mathbf{C}_{k-1}$ is the output of the previous multi-attention layer.

Positional encoding vectors are employed to encode the feature vector $F_k$. In the first Cross-attention sub-layer, where no preceding input exists, the query is learnable parameters $X$:

\begin{equation}
        \mathbfcal{A}\mathbf{ttn}(X,\mathbf{\Phi},\mathbf{\Phi}).   
\label{eq:cross_attention_0}
\end{equation}

\subsubsection{Self-Attention}

The self-attention sub-layer is placed right after the cross-attention sub-layer. Similar to~\cite{jaegle2021perceiver}, the goal is to model both short-range and long-range dependencies within the features and capture the global context. 

The inputs for the self-attention in layer $k$, denoted as $\mathbf{A}^S_{k}$, are the output of the cross-attention:

\begin{equation}
\mathbf{A}^S_{k}:=\mathbfcal{A}\mathbf{ttn}(\mathbf{A}^{C}_{k},\mathbf{A}^{C}_{k},\mathbf{A}^{C}_{k}).
\label{eq:self_attention}
\end{equation}

\subsubsection{View-Attention}
The view-attention sub-layer combines the latent features from the current view with the latent features from the other view, enabling the expansion of the context to both MLO and CC. The values $V$ of the view-attention block are the output of the preceding self-attention block, while the query $Q$ and keys $K$ are the output of the view-attention from the other branch at the corresponding level: 

\begin{equation}
\mathbfcal{A}\mathbf{ttn}(\hat{\mathbf{A}^S_{k}},\hat{\mathbf{A}^S_{k}},\mathbf{A}^S_{k}),
\label{eq:view_attention}
\end{equation}

where $\hat{\cdot}$ denotes the output of the self-attention sub-layer in the other branch.

\subsubsection{Input-Output}

To ensure that our multi-attention layer receives input with the same number of channels, we reshape the feature tensor $\mathbf{F}_k$ to have dimensions $\mathbf{F}_k \in \mathbb{R}^{d' \times N_k}$, where $N_k = \frac{H' \cdot W'}{2^{(n-k-1)}}$ and $d' = 4 \cdot d_0$ represents the desired length of the feature vectors inserted into the multi-attention layer.

The output tensors of the multi-attention layer at level $k$ have dimensions $\mathbb{R}^{L \times N_k}$, where $L$ denotes the length of the multi-attention latent vector. The query parameters $\mathbf{\hat{X}}$ and $\mathbf{\hat{Y}}$ are learnable parameters with dimensions $\mathbb{R}^{L \times N_Q}$, where $N_Q$ is a hyperparameter set by the user.

\subsection{Sub-Layer Attention Computation}

Given an input sequence $X = (x_1, x_2, \ldots, x_N)$, each attention sub-layer computes a weighted sum of the values at all positions in the sequence. This is achieved through the following steps:

\textbf{Positional Encoding}. To provide positional information to the model, we apply a Fourier feature encoding to the input sequences. Similar to \cite{jaegle2021perceiver}, we utilize the Fourier feature positional encodings introduced in \cite{tancik2020fourier}. 

Given $N$ input vectors $x_i \in \mathbb{R}^{L}$ each associated with a position index $i$, we first normalize the index as:

\begin{equation}
p_i = 2 \cdot \frac{i}{N} - 1  
\end{equation}

We then define a set of sinusoidal frequency bands: 

\begin{equation}
S_b = \{S \mid S = b \cdot \frac{m_{\text{freq}}}{n_{\text{bands}}}, 1 \leq b < n_{\text{bands}}, b \in \mathbb{Z}^{+}\}
\end{equation}

where $m_{\text{freq}}$ and $n_{\text{bands}}$ determine the maximum frequency and number of bands.

The sinusoidal position encoding vectors are then calculated as: 

\begin{equation}
PE1_b(p_i) = \sin(p_i \cdot S_b \cdot \pi),
\end{equation}

\begin{equation} 
PE2_b(p_i) = \cos(p_i \cdot S_b \cdot \pi).
\end{equation}

Finally, we concatenate the normalized index $p_i$ and encoding vectors $PE1_b$, $PE2_b$ to the original input $x_i$, expanding it to $x_i \in \mathbb{R}^{L + 2n_{\text{bands}} + 1}$. This injects positional information through sinusoidal functions of different frequencies, allowing the model to utilize the order of the input vectors.

\textbf{Linear Transformation}. The input sequence $X$ is linearly projected into the query ($Q$), key ($K$), and value ($V$) matrices using learnable weight matrices $W_Q$, $W_K$, and $W_V$:

\begin{align}
Q &= XW_Q \\
K &= XW_K \\  
V &= XW_V
\end{align}

where $Q, K, V \in \mathbb{R}^{L \times d_{\text{model}}}$, and $d_{\text{model}}$ is the dimensionality of the model. 

This transformation projects the input into distinct query, key, and value spaces. The query and key matrices are used to compute attention weights indicating the relevance between inputs. The value matrix holds the input representations that will be aggregated according to the attention weights.
  
\textbf{Attention Unit}. We compute the attention function (\Cref{eq:attention}).

\textbf{Position-wise Feed-Forward Network}. After the attention unit, a position-wise feed-forward network is applied to each position independently. The feed-forward network consists of two linear transformations with a ReLU activation function in between:

\begin{equation}
\text{FFN}(x) = \max(0, xW_1 + b_1)W_2 + b_2
\end{equation}

where $x$ is the input, $W_1$, $W_2$ are weight matrices, and $b_1$, $b_2$ are bias vectors.

\section{Experimental Setup}

\subsection{Dataset}

We use the Curated Breast Imaging Subset of DDSM (CBIS-DDSM) dataset~\cite{lee2017curated} that contains valuable metadata providing additional clinical information about each mammogram and associated lesions. It is a widely used mammography image collection annotated by radiologists, derived from the original DDSM~\cite{heath1998current} dataset and contains a diverse range of breast abnormalities, including benign and malignant lesions. The images are provided in the Digital Imaging and Communications in Medicine (DICOM) format, along with detailed annotation files. These files specify lesion locations, types, ROI crops, and binary masks across the craniocaudal (CC) and mediolateral oblique (MLO) views. It includes 1566 patients with total of 3,568 abnormalities, 1696 mass and 1872 calcification. In our experiments, we employ five-fold stratified cross-validation to maintain class balance across folds.

\subsection{Implementation Details}  

The training was done in mini-batches, with each mini-batch size set to $16$. For each image in our training data, the image's content is scaled to $1024\times1024$ pixels. As data augmentation, we used vertical and horizontal flips, as well as elastic deformation. We set a total of $1000$ training iterations for each fold. We utilized Cross-Entropy as our loss ~\cite{de2005tutorial}. An initial learning rate was set to 0.001 and we employed a decaying factor of $10$ after $500$ iterations. We used the SGD optimizer~\cite{robbins1951stochastic} with momentum set to $0.9$. Dropout value was set to $0.25$. We implemented our model in PyTorch~\cite{NEURIPS2019_9015}.

\begin{table*}[tb]
\centering
\begin{tabular}{ccccccc}
\toprule
\textbf{Model} & \textbf{AUC} & \textbf{Accuracy} & \textbf{Specificity} & \textbf{Precision} & \textbf{Recall} & \textbf{F1-Score} \\
\midrule
 \cite{mo2023hover} &0.680 & 0.661 & 0.670& 0.638 & 0.651 & 0.644 \\
 \cite{tulder2021multi} & 0.811 &     0.723 &        0.750 &      0.686 &   0.698 &     0.692 \\
 Ours - no descriptors &  0.711 & 0.664 & 0.650 & 0.676 & 0.619 &    0.634  \\
 Ours  &  0.872 & 0.760 & 0.773 & 0.760 & 0.743 & 0.751 \\
\bottomrule
\end{tabular}
\caption{Quantitative performance analysis to detect abnormality using CBIS-DDSM dataset.}
\label{table:results}
\end{table*}

\begin{table*}[tb]
\centering
\begin{tabular}{cccc|cccccc}
\toprule
Configuration & Q  & K  & V  &   AUC &  Accuracy &  Specificity &  Precision &  Recall &  F1-Score \\
\midrule
0 &  C &  O &  O & 0.835 &     0.738 &        0.731 &      0.643 &   0.750 &     0.692 \\
1 &  O &  C &  O & 0.850 &     0.760 &        0.790 &      0.762 &   0.727 &     0.744 \\
2 &  O &  O &  C & \textbf{0.878} & \textbf{0.796 }& 0.773 & \textbf{0.816} & \textbf{0.786 }&    \textbf{0.780} \\
3 &  C &  C &  O & 0.852 &     0.760 &        0.768 &      0.714 &   0.750 &     0.732 \\
4 &  C &  O &  C & 0.843 &     0.764 &        0.792 &      0.762 &   0.733 &     0.747 \\
5 &  O &  C &  C & 0.848 &     0.771 &       \textbf{ 0.803 }&      0.778 &   0.737 &     0.757 \\
\bottomrule
\end{tabular}\\ 
\caption{The effect of different configurations for the inputs (Query, Keys, Value) to the view attention sub-layer in the multi-attention layer. The inputs can come from either the current view (C) or the opposite view (O) of the mammogram. Configuration 2, in which the Query and Keys inputs are from the opposite view, achieved the best overall performance.}
\label{table:results_kqv}
\end{table*}

\section{Results}
We compare our multi-modal descriptor-based model ("Deep BI-RADS") against several baselines: a descriptor-excluded variant of our own model, a multi-view Transformer baseline~\cite{tulder2021multi}, and an advanced recent single-view Transformer approach with four branches~\cite{mo2023hover}. The descriptor-excluded variant includes the multi-attention layers, allowing us to evaluate the specific contribution of the BI-RADS descriptors. Comparing with a multi-view architecture helps assess the contribution of both the attention layers and the BI-RADS descriptors. The multi-view baseline employs a Transformer architecture to analyze pairs of unregistered mammograms from different views and achieves state-of-the-art results on the CBIS-DDSM. Comparison with a single-view Transformer evaluates the contribution of the multi-view architecture.

To ensure a fair comparison, we trained all models from scratch following their respective provided training protocols. We evaluate the models for classifying mass lesions, following common practice. The results are obtained using five-fold stratified cross-validation to maintain class balance across folds.

Table~\ref{table:results} presents the quantitative performance analysis. Our multi-modal approach achieves a higher AUC of 0.872 compared to 0.711 without incorporating the BI-RADS descriptors, demonstrating the benefits of integrating textual information. We also attain an AUC of 0.872 versus 0.811 for the baseline multi-view model, showcasing the advantages of our multi-attention fusion approach over prior multi-view only techniques.

Beyond AUC, utilizing BI-RADS descriptors enables consistent gains across accuracy, specificity, precision, recall, and F1-score on both tasks. Our approach increases recall from 0.619 to 0.743 compared to the baseline without BI-RADS. This demonstrates improved sensitivity in detecting true positive cases by incorporating textual descriptor classes.

Notably, the high F1 scores demonstrate that our model balances improved sensitivity with precision, rather than sacrificing one metric for the other. This indicates that our multi-modal methodology incorporates the radiologist context to enhance interpretation without introducing additional false positives.

\begin{figure}[!htb]
\centering
\includegraphics[width=0.65\linewidth]{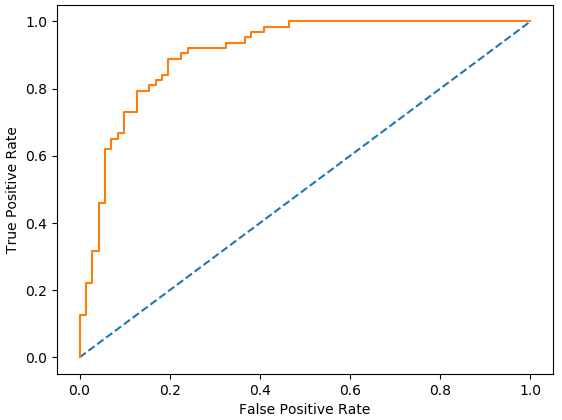}
\caption{ROC curve for our approach}
\label{fig:roc}
\end{figure}

\Cref{fig:roc} presents the ROC curve for a single fold, summarizing the trade-off between the true positive rate and false positive rate for our model using different probability thresholds.
Overall, our multi-modal method shows promise for generalized breast abnormality detection by effectively combining visual and textual information.

\subsection{Input to Multi-Attention Layer}

\begin{figure}[tb]
\centering
\includegraphics[width=0.9\textwidth]{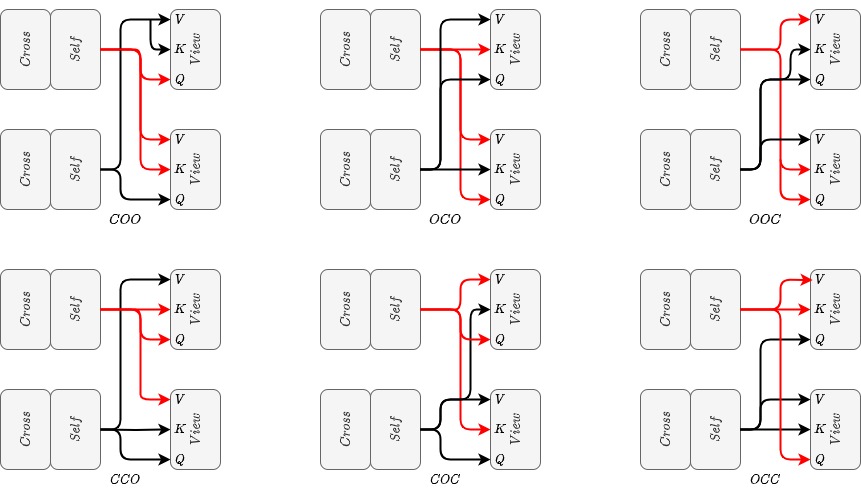}
\caption{The possible configurations of the inputs for the view attention sublayer in our multi-attention layer. Q, K, and V represent the Query, Keys, and Value respectively.}
\label{fig:ablation-connect}
\end{figure}

\Cref{table:results_kqv} represents different configurations (\Cref{fig:ablation-connect}) for wiring the query (Q), and keys (K), and values (V). In both branches, the input to the multi-attention layer can be either from the current view (C) or from the opposite view (O). There are six possible configurations, as we always wire at least one input from the other view. Based on the results in \Cref{table:results_kqv}, we conclude that wiring the query (Q) and keys (K) inputs to the attention layer from the opposite view (configuration 2) leads to the best performance, with the highest metrics. 

Some observations:

\begin{itemize}
    \item Wiring Q and K from the opposite view consistently outperforms wiring them from the current view (e.g., compare configurations 2 vs 3). This might suggest that the attention mechanism benefits from fusing information between the two views via the value input specifically.

    \item Based on how the inputs are interconnected between the two views, there is a noticeable difference in performance. This emphasizes the importance of effectively leveraging the two views.

    \item Wiring Q and K from the same view (configurations 2,3) performs better than wiring them from different views. 

    \item Specificity is highest when wiring Q from the opposite view and K and V from the current view (configuration 5). However, other metrics like recall are lower in this configuration.
    
\end{itemize}

\subsection{Number of Multi-Attention Layers}

The number of multi-attention layers primarily influences our model size. We trained configurations with 3, 5, 6, and 7 layers.

\Cref{table:perv_blocks} presents the accuracy for each model across five folds. The 3-layer model underperformed, while the 5-layer model achieved the second-best results overall. The 6-layer configuration yielded the highest average accuracy, outperforming 5 layers. However, further increasing layers to 7 degraded performance, likely due to overfitting given the limited dataset size. In our implementation, we deploy 6  layers which achieved the optimal trade-off between model capacity and overfitting on this dataset.

\begin{table}[tb]
    \centering
    \begin{tabular}{cc}
        \toprule
       \textbf{Configuration} &  \textbf{Average} \\
        \midrule
        3 layers & 0.69 $\pm$ 0.015 \\
        5 layers & 0.73 $\pm$ 0.019\\
        6 layers & 0.76 $\pm$ 0.010\\ 
        7 layers & 0.67 $\pm$ 0.018 \\
        \bottomrule
    \end{tabular}
    \caption{Accuracy for different numbers of  multi-attention layers. Results are across five folds.}
    \label{table:perv_blocks}
\end{table}

\subsection{Augmentations}
We tested various data augmentation strategies to improve model generalization of our model. \Cref{table:augs} presents test set performance for different augmentation configurations. The baseline with no augmentations underperformed all augmented models, indicating augmentations are beneficial. Adding random horizontal/vertical flips or elastic deformations to the baseline improved average accuracy. Resizing the images to 1024x1024 pixels achieved the best overall results. Interpolation sizes of 2048x2048 and 384x384 underperformed. Gaussian noise augmentation degraded performance, likely due to occluding meaningful mammographic details. The optimal configuration utilized interpolation upsampling to 1024x1024 pixels, which seems to balance overfitting and underfitting effects based on model capacity.

\begin{table}[tb]
    \centering
    \begin{tabular}{ll}
        \toprule
        \textbf{Configuration} & \textbf{Average} \\
        \midrule
        Baseline w/o aug.& 0.656 $\pm$ 0.025 \\
        Baseline + 384  &0.702  $\pm$ 0.014\\
        Baseline + 1024 & 0.72 $\pm$  0.008\\
        Baseline + 2048 & 0.646 $\pm$ 0.019\\ 
        Baseline + h/vflip & 0.666 $\pm$ 0.02\\
        Baseline + elastic & 0.688  $\pm$ 0.017\\
        Baseline + Gaussian & 0.612 $\pm$ 0.018\\
        \bottomrule
    \end{tabular}
    \caption{Accuracy for different augmentation strategies.}
    \label{table:augs}
\end{table}

\section{Conclusion}
In this study we ask whether incorporating BI-RADS descriptors can improve deep learning for cancer detection. Our results provide a clear affirmative answer to this question. We presented a multi-modal approach that combines visual mammogram data with textual BI-RADS descriptors, utilizing a dual-branch architecture with iterative attention layers. Experiments on the CBIS-DDSM dataset demonstrated significant improvements over image-only models. These findings suggests that the fusion of features based on human expertise and automatically extracted features can lead to superior outcomes in cancer detection.
%
%
%
%

\bibliographystyle{splncs04} 
\bibliography{references}

\begin{thebibliography}{10}
\providecommand{\url}[1]{\texttt{#1}}
\providecommand{\urlprefix}{URL }
\providecommand{\doi}[1]{https://doi.org/#1}

\bibitem{ACR-BIRADS}
{American College of Radiology}: ACR BI-RADS® Atlas — Mammography. American College of Radiology, Reston, VA (2013)

\bibitem{ACR-TIRADS}
{American College of Radiology}: ACR TI-RADS® Atlas. American College of Radiology, Reston, VA (2073)

\bibitem{DBLP:journals/corr/BaKH16}
Ba, L.J., Kiros, J.R., Hinton, G.E.: Layer normalization. CoRR  \textbf{abs/1607.06450} (2016), \url{http://arxiv.org/abs/1607.06450}

\bibitem{chen2022multi}
Chen, Y., Wang, H., Wang, C., Tian, Y., Liu, F., Liu, Y., et~al.: Multi-view local co-occurrence and global consistency learning improve mammogram classification generalisation. In: International Conference on Medical Image Computing and Computer-Assisted Intervention. pp. 3--13. Springer Nature Switzerland (2022)

\bibitem{de2005tutorial}
De~Boer, P.T., Kroese, D.P., Mannor, S., Rubinstein, R.Y.: A tutorial on the cross-entropy method. Annals of operations research  \textbf{134}(1),  19--67 (2005)

\bibitem{Aguilar2020}
Falconi, L.G., Maria~Perez, W.G.A., Conci, A.: Transfer learning and fine tuning in breast mammogram abnormalities classification on cbis-ddsm database  \textbf{5}(2),  154--165

\bibitem{he2016deep}
He, K., Zhang, X., Ren, S., Sun, J.: Deep residual learning for image recognition. In: Proceedings of the IEEE conference on computer vision and pattern recognition. pp. 770--778 (2016)

\bibitem{he2016identity}
He, K., Zhang, X., Ren, S., Sun, J.: Identity mappings in deep residual networks (2016), \url{http://arxiv.org/abs/1603.05027}, cite arxiv:1603.05027Comment: ECCV 2016 camera-ready

\bibitem{heath1998current}
Heath, M., Bowyer, K., Kopans, D., Kegelmeyer, P., Moore, R., Chang, K., Munishkumaran, S.: Current status of the digital database for screening mammography. In: Digital mammography, pp. 457--460. Springer (1998)

\bibitem{jaegle2021perceiver}
Jaegle, A., Gimeno, F., Brock, A., Vinyals, O., Zisserman, A., Carreira, J.: Perceiver: General perception with iterative attention. In: International conference on machine learning. pp. 4651--4664. PMLR (2021)

\bibitem{kolesnikov2020big}
Kolesnikov, A., Beyer, L., Zhai, X., Puigcerver, J., Yung, J., Gelly, S., Houlsby, N.: Big transfer (bit): General visual representation learning. In: Computer Vision--ECCV 2020: 16th European Conference, Glasgow, UK, August 23--28, 2020, Proceedings, Part V 16. pp. 491--507. Springer (2020)

\bibitem{lee2017curated}
Lee, R.S., Gimenez, F., Hoogi, A., Miyake, K.K., Gorovoy, M., Rubin, D.L.: A curated mammography data set for use in computer-aided detection and diagnosis research. Scientific data  \textbf{4}(1), ~1--9 (2017)

\bibitem{liu2021deep}
Liu, H., Chen, Y., Zhang, Y., Wang, L., Luo, R., Wu, H., Wu, C., Zhang, H., Tan, W., Yin, H., et~al.: A deep learning model integrating mammography and clinical factors facilitates the malignancy prediction of bi-rads 4 microcalcifications in breast cancer screening. European Radiology  \textbf{31},  5902--5912 (2021)

\bibitem{liu2020cross}
Liu, Y., Zhang, F., Zhang, Q., Wang, S., Wang, Y., Yu, Y.: Cross-view correspondence reasoning based on bipartite graph convolutional network for mammogram mass detection. In: 2020 IEEE/CVF Conference on Computer Vision and Pattern Recognition (CVPR). pp. 3811--3821 (2020)

\bibitem{mo2023hover}
Mo, Y., Han, C., Liu, Y., Liu, M., Shi, Z., Lin, J., Zhao, B., Huang, C., Qiu, B., Cui, Y., et~al.: Hover-trans: Anatomy-aware hover-transformer for roi-free breast cancer diagnosis in ultrasound images. IEEE Transactions on Medical Imaging  (2023)

\bibitem{9871564}
Nguyen, H.T.X., Tran, S.B., Nguyen, D.B., Pham, H.H., Nguyen, H.Q.: A novel multi-view deep learning approach for bi-rads and density assessment of mammograms. In: 2022 44th Annual International Conference of the IEEE Engineering in Medicine and Biology Society (EMBC). pp. 2144--2148 (2022). \doi{10.1109/EMBC48229.2022.9871564}

\bibitem{NEURIPS2019_9015}
Paszke, A., Gross, S., Massa, F., Lerer, A., Bradbury, J., Chanan, G., Killeen, T., Lin, Z., Gimelshein, N., Antiga, L., Desmaison, A., Kopf, A., Yang, E., DeVito, Z., Raison, M., Tejani, A., Chilamkurthy, S., Steiner, B., Fang, L., Bai, J., Chintala, S.: Pytorch: An imperative style, high-performance deep learning library. In: Wallach, H., Larochelle, H., Beygelzimer, A., d\textquotesingle Alch\'{e}-Buc, F., Fox, E., Garnett, R. (eds.) Advances in Neural Information Processing Systems 32, pp. 8024--8035. Curran Associates, Inc. (2019)

\bibitem{journals/corr/abs-1903-10520}
Qiao, S., Wang, H., Liu, C., Shen, W., Yuille, A.L.: Weight standardization. CoRR  \textbf{abs/1903.10520} (2019), \url{http://dblp.uni-trier.de/db/journals/corr/corr1903.html#abs-1903-10520}

\bibitem{robbins1951stochastic}
Robbins, H., Monro, S.: A stochastic approximation method. The annals of mathematical statistics pp. 400--407 (1951)

\bibitem{shen2021interpretable}
Shen, Y., Wu, N., Phang, J., Park, J.C., Liu, K., Tyagi, S., et~al.: An interpretable classifier for high-resolution breast cancer screening images utilizing weakly supervised localization. Medical Image Analysis  \textbf{68},  101908 (2021)

\bibitem{tancik2020fourier}
Tancik, M., Srinivasan, P., Mildenhall, B., Fridovich-Keil, S., Raghavan, N., Singhal, U., Ramamoorthi, R., Barron, J., Ng, R.: Fourier features let networks learn high frequency functions in low dimensional domains. Advances in Neural Information Processing Systems  \textbf{33},  7537--7547 (2020)

\bibitem{tsai2022high}
Tsai, K.J., Chou, M.C., Li, H.M., Liu, S.T., Hsu, J.H., Yeh, W.C., Hung, C.M., Yeh, C.Y., Hwang, S.H.: A high-performance deep neural network model for bi-rads classification of screening mammography. Sensors  \textbf{22}(3), ~1160 (2022)

\bibitem{tulder2021multi}
Tulder, G.v., Tong, Y., Marchiori, E.: Multi-view analysis of unregistered medical images using cross-view transformers. In: International Conference on Medical Image Computing and Computer-Assisted Intervention. pp. 104--113. Springer (2021)

\bibitem{tulder2021cross}
van Tulder, G., Tong, Y., Marchiori, E.: Multi-view analysis of unregistered medical images using cross-view transformers. In: International Conference on Medical Image Computing and Computer-Assisted Intervention. pp. 104--113. Springer (2021)

\bibitem{vaswani2017attention}
Vaswani, A., Shazeer, N., Parmar, N., Uszkoreit, J., Jones, L., Gomez, A.N., Kaiser, L., Polosukhin, I.: Attention is all you need. Advances in Neural Information Processing Systems  \textbf{30},  5998--6008 (2017)

\bibitem{Veeling2018-qh}
Veeling, B.S., Linmans, J., Winkens, J., Cohen, T., Welling, M.: Rotation equivariant {CNNs} for digital pathology  (Jun 2018)

\bibitem{DBLP:conf/eccv/WuH18}
Wu, Y., He, K.: Group normalization. In: Ferrari, V., Hebert, M., Sminchisescu, C., Weiss, Y. (eds.) Computer Vision - {ECCV} 2018 - 15th European Conference, Munich, Germany, September 8-14, 2018, Proceedings, Part {XIII}. Lecture Notes in Computer Science, vol. 11217, pp. 3--19. Springer (2018). \doi{10.1007/978-3-030-01261-8\_1}, \url{https://doi.org/10.1007/978-3-030-01261-8\_1}

\bibitem{yan2021towards}
Yan, Y., Conze, P.H., Lamard, M., Quellec, G., Cochener, B., Coatrieux, G.: Towards improved breast mass detection using dual-view mammogram matching. Medical Image Analysis  \textbf{71},  102083 (2021)

\bibitem{zhang2021bi}
Zhang, B., Vakanski, A., Xian, M.: Bi-rads-net: an explainable multitask learning approach for cancer diagnosis in breast ultrasound images. In: 2021 IEEE 31st International Workshop on Machine Learning for Signal Processing (MLSP). pp.~1--6. IEEE (2021)

\end{thebibliography}

\end{document}